\documentclass{article}

\usepackage{arxiv}

\usepackage{natbib}

\usepackage{algorithm}
\usepackage{algorithmic}
\usepackage{amsmath,graphicx}
\usepackage{amsfonts}
\usepackage{hyperref}
\usepackage{url}
\usepackage{bm}
\usepackage{xcolor}
\usepackage{adjustbox}
\usepackage{multirow,tabularx}

\usepackage{adjustbox}
\usepackage{subcaption,booktabs}

\newcommand\blfootnote[1]{%
  \begingroup
  \renewcommand\thefootnote{}\footnote{#1}%
  \addtocounter{footnote}{-1}%
  \endgroup
}

\def\R{{\mathbb R}}

\title{An End-to-End Time Series Model for Simultaneous Imputation and Forecast}

\author{
Trang H. Tran$^{1,2}$\thanks{Work done while an intern at IBM Research} \qquad 
Lam M. Nguyen$^{2}$ \qquad 
Kyongmin Yeo$^{2}$ \qquad
Nam Nguyen$^{2}$ \qquad 
Dzung Phan$^{2}$ \\
\textbf{Roman Vaculin}$^{2}$ \qquad
\textbf{Jayant Kalagnanam}$^{2}$ \\
$^{1}$ School of Operations Research and Information Engineering, Cornell University, Ithaca, NY, USA\\
$^{2}$ IBM Research, Thomas J. Watson Research Center, Yorktown Heights, NY, USA \\
\texttt{htt27@cornell.edu},
\texttt{LamNguyen.MLTD@ibm.com}
}

\begin{document}

\maketitle

\begin{abstract}
Time series forecasting using historical data has been an interesting and challenging topic, especially when the data is corrupted by missing values. In many industrial problem, it is important to learn the inference function between the auxiliary observations and target variables as it provides additional knowledge when the data is not fully observed. 
We develop an end-to-end time series model that aims to learn the such inference relation and make a multiple-step ahead forecast. Our framework trains jointly two neural networks, one to learn the feature-wise correlations and the other for the modeling of temporal behaviors. 
Our model is capable of simultaneously imputing the missing entries and making a multiple-step ahead prediction.
The experiments show good overall performance of our framework over existing methods in both imputation and forecasting tasks.
\end{abstract}

\blfootnote{Correspondence to: Lam M. Nguyen.}

\section{Introduction}

Learning the complex structure of multivariate time series has been one of the major interests across many application domains, including economics, transportation, manufacturing \citep{gpvae,autoformer,logtrans,informer}. While there has been much progress in the data-driven learning and processing complex time series, it still remains as a challenging topic, in particular, when the data is corrupted \citep{brits,effects,mrnn,saits}. In this paper, we consider the forecasting task which aims to make prediction of future values  using historical data that may contain missing values.

In addition, for many industrial problems, the time series features can be in two categories: auxiliary features ($\bm{X}$) that provide information about the state of a system and target variables ($\bm{Y}$) that depends on the auxiliary features and may convey valuable information. For example, in the operation of a chemical reactor, the auxiliary features include temperature, pressure and concentration of chemicals observed through a sensor network, while the target variable may include the quality of the material and throughput. We are interested in the time series problem where the data set consists of $\bm{X}$ and $\bm{Y}$. 
In general, $\bm{X}$ is more readily available, as it is obtained from a sensor network, while $\bm{Y}$ may be temporally sparse since it may be expensive or difficult to collect the data. This so-called soft sensor problem has been of interest in many industrial applications \citep{softsensor1,softsensor3}.

In a formal setting, we consider a time series data, which consists of two sets of variables $\bm{X}=\{\bm{x}_1,\cdots,\bm{x}_T\}$ and $\bm{Y} = \{\bm{y}_1,\cdots,\bm{y}_T\}$. We assume that $\bm{Y}$ can be inferred from $\bm{X}$, i.e., there exists a map, $f:\bm{X} \rightarrow \bm{Y}$. It is desired to learn the inference function $f$ based on historical time series data, when there are missing entries in both $\bm{X}$ and $\bm{Y}$. Moreover, we aim to forecast the future state of the system. While the conventional workflow of time series analysis with missing data consists of two sequential steps, \emph{i.e.}, imputation then training, we aim to develop an end-to-end model, in which imputation and training are performed simultaneously.
We summarize our contribution as follows. 
\begin{itemize}
    \item We propose a time series model that simultaneously imputes the missing data entries and makes a multiple-step ahead prediction. The experiments show good overall performance of our model over state-of-the-art methods in both tasks.
    \item The proposed method aims to learn the mapping from the auxiliary features to the target features under missing entries, and use the information to predict the future system states. It is desirable to learn such feature inference since one could predict the desired features from a set of observations, and more importantly, from a set of observations with missing values. 
    \item While our framework have the flexibility to use various different architectures, we propose to use a specific structure of the spatial learner and temporal learner functions to regularized the model behavior. Our ablation study validates the choice of training with the losses from the temporal inference function and spatial inference function as it performs better than individual loss. 
\end{itemize}

\section{Literature Review}

Classical approaches for time series processing is using statistical models e.g. ARIMA, autoregressive integrated moving average \citep{arima} and exponential smoothing \citep{expsmoothing}. However, these methods often require domain expertise and careful feature engineering. More modern approaches involves the use of machine learning models with abundant data.  
With the advent of deep learning, recurrent neural networks (RNN) as a nonlinear state-space model have become popular \citep{rnn} due to its strength in learning nonlinear dynamics from data.
Recently, convolutional neural networks (CNNs) and attention based methods \citep{empirical, attention} have attracted attention in the time series modeling, due to their immense success in image processing and natural language processing, respectively. 
We outline the brief literature survey of machine learning models for imputation and forecasting below.

\textbf{Imputation Literature}.
Some prior work proposed to address the missing data problem with statistical approaches, however they may require strong assumptions on missing values. Conventional approach relies on classical auto-regressive models to fill in the missing values \citep{arima2, arima}, of which accuracy quickly degrades as the fraction of missing entries increases.
Other popular approaches include matrix completion or fitting a parametric data-generating model \citep{temporal, methods}, but it usually requires low-rank assumptions or prior knowledge on the data generating distribution.

To relax these limitations, deep learning models have been adopted. The RNN-based imputation models, such as M-RNN \citep{mrnn} and BRITS \citep{brits}, aims to learn the mapping from the internal states to the time series data. While BRITS treats the imputed values as variables of RNN graph, M-RNN designs them to be constants. In addition, M-RNN do not consider the correlations among features as in BRITS. 
GP-VAE \citep{gpvae} model use a variational auto-encoder (VAE) architecture for time series imputation with a Gaussian process (GP) in the latent space.
Recently, researchers have explored various attention architectures for time series. 
NRTSI \citep{nrtsi} deals with irregularly-sampled time series by using two-loop structrure and treating the data as set of (time, data) tuples. SAITS \citep{saits} aims to learn missing values from a weighted combination of two diagonally-masked self-attention blocks.

\textbf{Forecasting}.
There are two major approaches for multiple-step ahead forecasting. The first approach aims to compute the joint probability distribution of the future system states by iteratively computing the time evolution using, \emph{e.g.}, RNNs \citep{rnn, Yeo22}. 
The second approach involves training a time series model that can directly forecast the next time steps based on the input of historical data. We focus on this approach since it has demonstrated better prediction accuracy in various forecasting tasks \citep{autoformer}.

Since the success of attention-based models in natural language processing \citep{attention}, attention-based time series models have become popular. However, the quadratic time and memory complexity of the attention mechanism poses a challenge in learning a long-range temporal correlation. LogTrans \citep{logtrans} and Pyraformer \citep{pyraformer} propose to use sparsity bias and reduce the computational complexity of the models. Informer \citep{informer} and FEDformer \citep{fedformer} utilize the low-rank properties of the self-attention matrix to enhance the performance. 
Autoformer \citep{autoformer} designs a decomposition architecture with an auto-correlation mechanism instead of the traditional attention-based models. On the other hand, \citep{linearmodels} proposes to use a simple set of linear models and suggests that the simple models may outperform complex structures.

\textbf{Multivariate time series dynamic}. While the univariate time series processing only focuses on the temporal relation between the time steps, dealing with multivariate time series is more challenging since it involves various features in the spatial dimension \cite{spatiotemporal_jake}. Some recent works propose to use the spatial-temporal methods which typically involves signals between two points of different time steps and features in a graph neural network \cite{spatiotemporal1, spatiotemporal2, spatiotemporal_jake}. However, these approaches often model in high dimensional space (product of time $T $ and dimension $d$) and they are expensive. In this paper, we use a two-step framework that allows restriction on the dynamic while still offer flexibility to learn the feature relations. 

In the next section, we describe the problem setting and our training framework for time series with missing values. While our method is general, we provide guidance on using a specific structure of inference functions to effectively utilize our framework.

\section{An End-to-end Framework}
\subsection{Problem Setting}
Our framework has the ability to handle training data with various level of missing values. Note that practical time series datasets often have \textit{nan} indicating missing values and imputation models use a binary masked matrix which replaces the missing values by 0 \citep{brits,ZhangTime}. We use the same approach that pair our (missing) data with $\bm{D}$ a masked matrix $\bm{M}_D$ as the input of the inference functions.

Let the time series data at time step $i$ be $\bm{z}_i=\{\bm{x}_i,\bm{y}_i\} \in \mathbb{R}^{d}$, in which $\bm{x}_i \in \mathbb{R}^{d_x}$, $\bm{y}_i \in \mathbb{R}^{d_y}$ and $d = d_x + d_y$. Here, we consider the input-$I$-predict-$O$ setting, where the model has the information of previous $I$ time steps and aim to predict the next $O$ future time steps. Note that there are other approach (i.e. using auto-regression models) that iteratively predicts one-step ahead and construct further predictions. However, we consider the input-$I$-predict-$O$ setting because it is more favorable in  many machine learning settings.  

Therefore our input data contains the first $I$ time steps:  $\bm{X}  = \{\bm{x}_1,\cdots,\bm{x} \}$ and $\bm{Y} =\{\bm{y}_1,\cdots,\bm{y} \}$ where  $\bm{Z} =\{\bm{X} ,\bm{Y} \} \in \mathbb{R}^{I\times d}$. Similarly, we can define the prediction data which contains the next $O$ time steps: $\bm{X}_O = \{\bm{x}_{I+1},\cdots,\bm{x}_{I+O}\}$, $\bm{Y}_O = \{\bm{y}_{I+1},\cdots,\bm{y}_{I+O}\}$, and $\bm{Z}_O = \{\bm{X}_O,\bm{Y}_O\}\in \mathbb{R}^{O \times d}$. 
In addition, we denote  $\bm{Z}_{+} = 
    \begin{bmatrix}
    \bm{Z} \\ \bm{Z}_{O}
    \end{bmatrix} \in \mathbb{R}^{(I+O) \times d}$. 
This study consider the multiple-step ahead forecast problem i.e. predicting $\bm{Z}_O$ based on available information from the first $I$ time steps. 
Note that $\bm{Z} $ and $\bm{Z}_O$ is the ground truth data without any missing entries. Thus, we use the asterisk $^*$ to denote the respective observable data \emph{i.e. }$\bm{Z} ^*$ and $\bm{Z}_O^*$ that potentially contains missing entries.

\subsection{Model Structure}
\begin{figure}[hpt!]
    \centering
    \includegraphics[width = 0.9\textwidth]{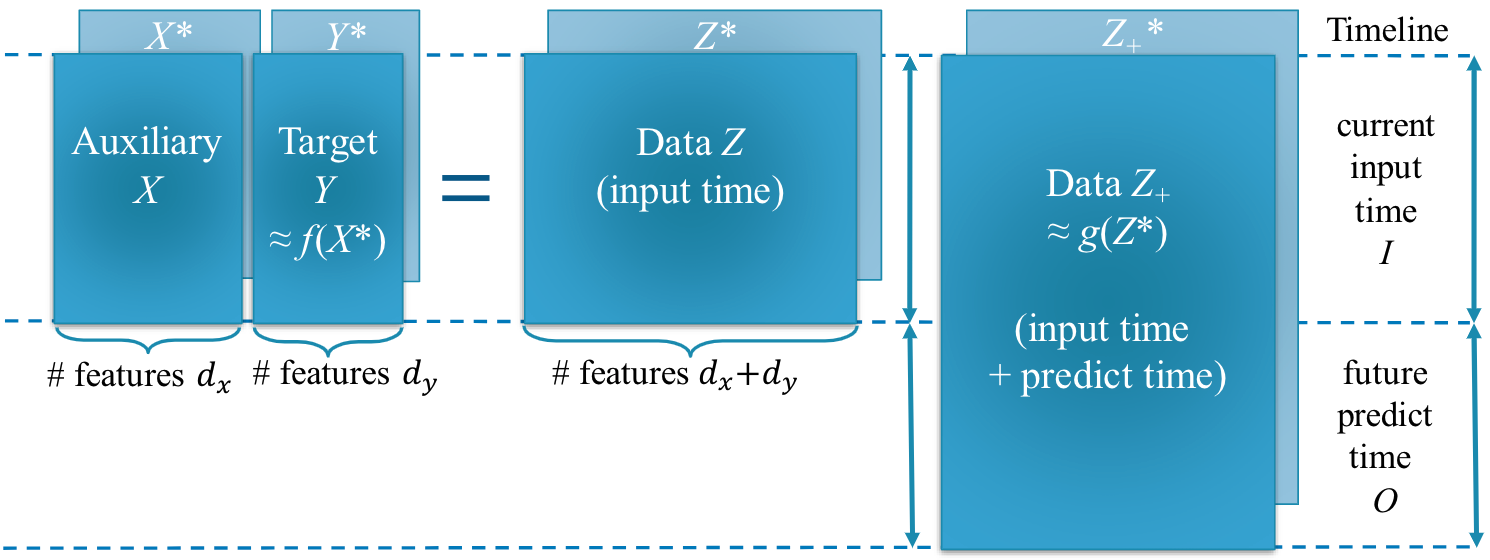}
    \caption{Description of learning dynamic system based on our Framework \ref{fig:fw}. The data $\bm{X}, \bm{Y}, \bm{Z}$ and $\bm{Z}_O$ denote the truth values while the observations $\bm{X}^*, \bm{Y}^*, \bm{Z}^*$ and $\bm{Z}_+^*$ have missing values. 
    }
    \label{fig:fw}
\end{figure}

First, we assume the time series data satisfies the following inference relation between the variables $\bm{X}$ and $\bm{Y}$: 
\begin{equation}
\bm{Y} = f(\bm{X}^*),
\end{equation}
where we denote $f$ as the spatial relation learner since it predicts the target features from the auxiliary features. Moreover, we aim to learn a function
\begin{align}
    \bm{Z}_{+} = g(\bm{Z}^*), \text{ where } \bm{Z}_{+} = 
    \begin{bmatrix}
    \bm{Z} \\ \bm{Z}_{O}
    \end{bmatrix} \in \mathbb{R}^{(I+O) \times d}.
\end{align}
where $g$ denotes the temporal relation learner that predicts the complete current and future data based on current inputs (potentially) with missing values. 
Figure 1 shows the dynamic of this time series problem. Our framework describes a generic dynamic system that allows the model to learn the spatial relation $f$ and the temporal relation $g$ simultaneously. The function $f$ and $g$ can be built as neural networks or any other approximation. Thus our learning model consists of these two functions.
Let $w_f$ and $w_g$ denotes the corresponding parameter of two learner functions $f$ and $g$, respectively. We propose the following loss function:
\begin{align*}
    L (\bm{Z}) &= \|(\bm{Y}^* - f (\bm{X}^*)) \odot \bm{M}_{Y^*}\|_F^2 +  \|(\bm{Z}_{+}^* - g (\{\bm{X}^*, f(\bm{X}^*)\}) \odot \bm{M}_{Z_{+}^*}\|_F^2 + r(w_f,w_g), 
\end{align*}
where $\bm{Z} = \{\bm{X}, \bm{Y}\}$;  
$\odot$ denotes the Hadamard product \emph{i.e. }element-wise multiplication and $r(w_f,w_g)$ is a regularization function on the weights. Due to space limit, we delay the description of the regularizer to the supplementary materials. Note that when the training data contains missing entries, we can only evaluate the loss function with observable data, that $\bm{M}_{Y^*}$ and $\bm{M}_{Z_{+}^*}$ are the masked matrices on the data $\bm{Y^*}$ and $\bm{Z}_{+}^*$ respectively.

The first term of the loss function aims to minimize the distance between the prediction $f(\bm{X}^*)$ and true target value $\bm{Y}$, and the second term aims to enforce the temporal relation between the first observed $I$ time steps (which may contain missing values) and the complete data for $I + O$ time steps. This approach offers the advantage of imputing current data and forecasting future data simultaneously: while the model $g$ strives to replicate the complete current data, it also uses the same information to predict future values. Intuitively, since the time stamps of  $\bm{Z}$ and $\bm{Z}_{O}$ are next to each other, they share some common characteristics. Thus, it is beneficial for the model to learn the complete data, as the predictor for current time may help regularize the predictor of future time, and vice versa. 

In the second loss function, we use the output $f(\bm{X}^*)$ of the spatial learner function $f$ instead of the value $\bm{Y}^*$. This choice helps the model learn to impute and forecast even when the information of the target feature $\bm{Y}^*$ is not available. In that setting, the prediction is made by two consecutive steps: (1) approximating $\bm{Y}$ by $f(\bm{X}^*)$ and (2) passing the concatenated data $\{\bm{X}^*, f(\bm{X}^*)\}$ through the function $g$ to obtain a prediction of $\bm{Z}_+$.
In the next section, we explain the principal ideas for the designs of $f$ and $g$, and we describe some network structures that utilize our framework. 
\subsection{Inference Functions for Framework \ref{fig:fw}}\label{subsec_learner}

Our framework design uses two inference functions to learn the time series dynamic, where $f$ and $g$ are responsible for the spatial relation and temporal relation, respectively. As their inputs are two-dimensional matrices, one can allow cross-relation between two entries of different time steps and features by flattening the matrices and building a network architecture on the flattened array. This is related to the spatial-temporal methods \cite{spatiotemporal1, spatiotemporal2, spatiotemporal_jake} which typically build graph neural networks to learn the time series dynamic. These neural networks often employ special and/or restricted connections because fully connected networks involve a lot of parameters and are expensive to learn. Our framework is flexible as one can design different architecture for the learning functions $f$ and $g$, and there is no restrictions on them. While there are different approaches for such design, we now demonstrate a specific neural network structure to validate our framework.

\textbf{Structure of Learner Functions}.
Since the two inference functions have a close connection with the spatial and temporal dynamic of the time series dataset, it is beneficial to restrict the temporal relation in the function $f$, and similarly the spatial relation in the function $g$. Then the cross-relations are learnt implicitly by the composite function $g (\{\bm{X}^*, f(\bm{X}^*)\} $ in our framework. 
Our ablation study in Section \ref{sec_ablation} validates that this architecture shows a better performance than the fully connected network approach which flattens the time and space dimensions.

 \begin{figure}[hpt!]
    \centering
    \includegraphics[width = 0.9\textwidth]{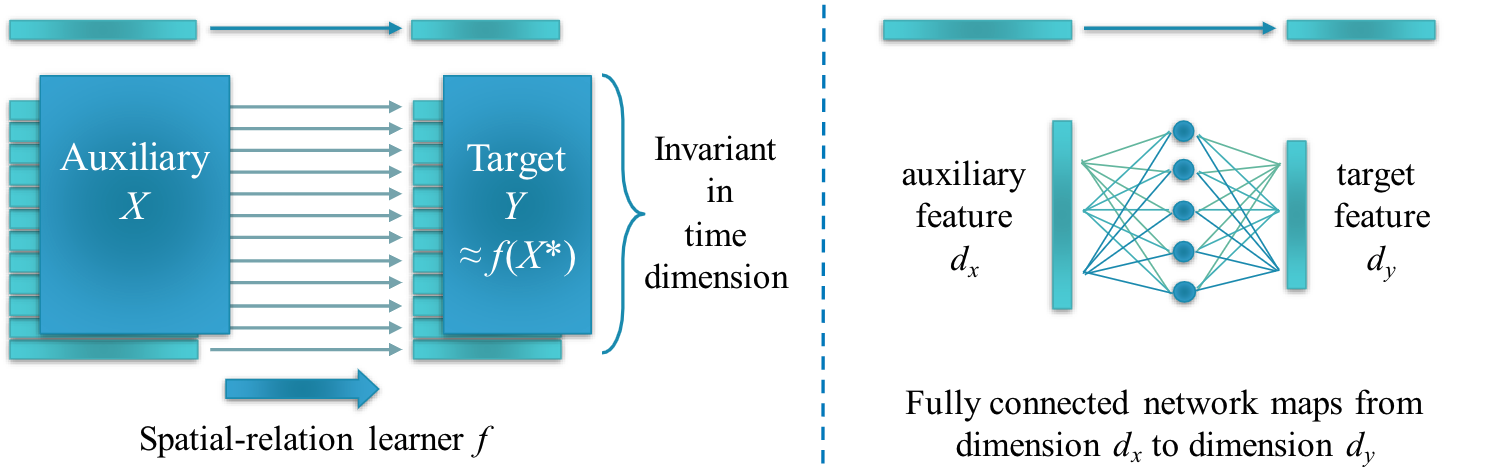}
    \caption{Description of the choice of spatial relation learner function $f$.}
    \label{fig:spatial}
\end{figure}
We describe the structure of the spatial relation function $f$ in Figure \ref{fig:spatial}. The design of the temporal relation function $g$ is similar. 
Firstly we observe that $f$ has input $\bm{X}^* \in \R^{I \times d_x}$ and output $\bm{Y} \in \R^{I \times d_y}$. Thus the input and output share the same time dimension with size $I$. 
For this reason, we let $f$ broadcasts a function $\hat{f}$ from dimension $d_x$ to dimension $d_y$: $\hat{f}(\bm{x}^*_i) = \bm{y_i}$ for every time steps $i$. Hence all the time step vectors share a common function $\hat{f}$, as in the left panel of  Figure \ref{fig:spatial}. This choice implies that the inference function is invariant in time dimension, thus allows it to utilize learning the relations between the auxiliary variable $x$ and the target variable $y$. 

In a similar manner, the temporal relation $g$ has input $\bm{Z}^* \in \R^{I \times d}$ and output $Z_{+}\in \R^{(I+O) \times d}$, in which $d = d_x+d_y$. Here the input and output share the space dimension, we let $g$ broadcasts a function $\hat{g}$ from dimension $I$ to dimension $I + O$. The function $g$ supports the temporal relation between current and future time steps based on observed current data. 

In Section \ref{sec_exp}, we demonstrate the effectiveness of our framework and the choice of the inference functions on various imputation and forecasting tasks. 
In all the experiments, we implement simple fully connected neural networks for $\hat{f}$ and $\hat{g}$ (e.g. the right panel of  Figure \ref{fig:spatial}). The number of parameters of our model is light (as it scales linearly with the sum of the input dimensions).  Due to space limit, we delay further detailed implementations to the Appendix. 

\section{Experiments}\label{sec_exp}

\subsection{Imputation Task}
\textbf{Settings.} 
In this experiment, we compare our method with two na\"{i}ve imputation method (Last and Median), and five state-of-the-art models: 
M-RNN \citep{mrnn}; GP-VAE \citep{gpvae}; BRITS \citep{brits}, Transformer \citep{attention} and SAITS \citep{saits}. 
We use Electricity Load Diagrams data set from UCI \citep{uci} for our imputation experiment. The data describes electricity consumption in kWh in every 15 minutes for 370 clients, with no missing data. The dataset spans for four years from 2011 to 2014. 
In order to make fair comparisons, we replicate the exact setting of the SAITS paper \citep{saits} with the same test set, validation set and training set. A sample of the model input has 100 time steps. 

\setlength{\tabcolsep}{1pt}
\setlength\doublerulesep{0.8pt}
\renewcommand{\arraystretch}{1.1}
\begin{table}[ht]
\caption{Comparison of the imputation accuracy for the Electricity data. The level of the missing data varies from 20\% to 90\%. 
We highlight {\color{red}\textbf{the best}} results and {\color{blue}\underline{the second best}} results.
}
\label{tab:my-table3}
\begin{center}
\begin{tabular}{c*{4}{ccc}}
\toprule[1pt]\midrule[0.3pt]
{Measures} & {MAE} & {RMSE} & {MRE} & {MAE} & {RMSE} & {MRE} & {MAE} & {RMSE} & {MRE} & {MAE} & {RMSE} & {MRE} \\
Missing \% & \multicolumn{3}{c}{20\%} & \multicolumn{3}{c}{30\%} & \multicolumn{3}{c}{40\%} & \multicolumn{3}{c}{50\%} \\
\hline
{\color{red}\textbf{Our method}} & {\color{red}\textbf{0.194}} & {\color{red}\textbf{0.288}} & {\color{red}\textbf{10.4\%}} & {\color{red}\textbf{0.249}} & {\color{red}\textbf{0.373}} & {\color{red}\textbf{13.3\%}} & {\color{red}\textbf{0.393}} & {\color{red}\textbf{0.561}} & {\color{red}\textbf{21.0\%}} & {\color{red}\textbf{0.537}} & {\color{red}\textbf{0.751}} & {\color{red}\textbf{28.7\%}} \\
SAITS & {\color{blue}\underline{0.763}} & {\color{blue}\underline{1.187}} & {\color{blue}\underline{40.8\%}} & {\color{blue}\underline{0.790}} & {\color{blue}\underline{1.223}} & {\color{blue}\underline{42.3\%}} & {\color{blue}\underline{0.869}} & {\color{blue}\underline{1.314}} & {\color{blue}\underline{46.7\%}} & {\color{blue}\underline{0.876}} & {\color{blue}\underline{1.377}} & {\color{blue}\underline{46.9\%}} \\
Transformer & 0.843 & 1.318 & 45.1\% & 0.846 & 1.321 & 45.3\% & 0.876 & 1.387 & 46.9\% & 0.895 & 1.410 & 47.9\% \\
BRITS & 0.928 & 1.395 & 49.7\% & 0.943 & 1.435 & 50.4\% & 0.996 & 1.504 & 53.4\% & 1.037 & 1.538 & 55.5\% \\
GP-VAE & 1.124 & 1.502 & 60.2\% & 1.057 & 1.571 & 56.6\% & 1.090 & 1.578 & 58.4\% & 1.097 & 1.572 & 58.8\% \\
M-RNN & 1.242 & 1.854 & 66.5\% & 1.258 & 1.876 & 67.3\% & 1.269 & 1.884 & 68.0\% & 1.283 & 1.902 & 68.7\% \\
Last & 1.012 & 1.547 & 54.2\% & 1.018 & 1.559 & 54.5\% & 1.025 & 1.578 & 54.9\% & 1.032 & 1.595 & 55.2\% \\
Median & 2.053 & 2.726 & 109.9\% & 2.055 & 2.732 & 110.0\% & 2.058 & 2.734 & 110.2\% & 2.053 & 2.728 & 109.9\% \\
\hline
\midrule[0.3pt]
Missing \% & \multicolumn{3}{c}{60\%} & \multicolumn{3}{c}{70\%} & \multicolumn{3}{c}{80\%} & \multicolumn{3}{c}{90\%} \\
\hline
{\color{red}\textbf{Our method}} & {\color{red}\textbf{0.643}} & {\color{red}\textbf{0.885}} & {\color{red}\textbf{34.4\%}} & {\color{red}\textbf{0.837}} & {\color{red}\textbf{1.134}} & {\color{red}\textbf{44.8\%}} & {\color{red}\textbf{0.903}} & {\color{red}\textbf{1.208}} & {\color{red}\textbf{48.4\%}} & 0.969 & {\color{red}\textbf{1.296}} & 51.9\% \\
SAITS & 0.892 & {\color{blue}\underline{1.328}} & 47.9\% & {\color{blue}\underline{0.898}} & {\color{blue}\underline{1.273}} & {\color{blue}\underline{48.1\%}} & {\color{blue}\underline{0.908}} & {\color{blue}\underline{1.327}} & {\color{blue}\underline{48.6\%}} & {\color{red}\textbf{0.933}} & {\color{blue}\underline{1.354}} & {\color{blue}\underline{49.9\%}} \\
Transformer & {\color{blue}\underline{0.891}} & 1.404 & {\color{blue}\underline{47.7\%}} & 0.920 & 1.437 & 49.3\% & 0.924 & 1.472 & 49.5\% & {\color{blue}\underline{0.934}} & 1.491 & {\color{red}\textbf{49.8\%}} \\
BRITS & 1.101 & 1.602 & 59.0\% & 1.090 & 1.617 & 58.4\% & 1.138 & 1.665 & 61.0\% & 1.163 & 1.702 & 62.3\% \\
GP-VAE & 1.101 & 1.616 & 59.0\% & 1.037 & 1.598 & 55.6\% & 1.062 & 1.621 & 56.8\% & 1.004 & 1.622 & 53.7\% \\
M-RNN & 1.298 & 1.912 & 69.4\% & 1.305 & 1.928 & 69.9\% & 1.318 & 1.951 & 70.5\% & 1.331 & 1.961 & 71.3\% \\
Last & 1.040 & 1.615 & 55.7\% & 1.049 & 1.640 & 56.2\% & 1.059 & 1.663 & 56.7\% & 1.070 & 1.690 & 57.3\% \\
Median & 2.057 & 2.734 & 110.2\% & 2.050 & 2.726 & 109.8\% & 2.059 & 2.734 & 110.2\% & 2.056 & 2.723 & 110.1\%\\
\midrule[0.3pt]\bottomrule[1pt]
\end{tabular}
\end{center}
\end{table}

For the purpose of the imputation experiment, we randomly remove 20 to 90 percent of the original data $\bm{Z}$ to test our imputation model. This practice is consistent with the  experiments for SAITS \citep{saits}, where the training model and validation step do not have access to the complete ground truth. We refer to this procedure as artificially masking. 
Note that the test results are computed on the complete test data $\bm{Z}$ (without artificially missing values) as the test model have access to the true data. We report the mean absolute error (MAE), root mean square error (RMSE), and mean relative error (MRE), where lower metrics demonstrate better results.
In this experiment, we set the last 50 features to be the features of interest $d_y$. This is consistent with the forecasting experiments in the next sections. The experiments are repeated five times, and and we report the average results. 

\textbf{Results.} 
Table \ref{tab:my-table3} shows the performance of our method comparing to several imputation benchmarks. 
The results for these methods are from the reference \citep{saits}, and they perform experiment with the data sets $\bm{Z}_*$ with varying level of missing entries. In Table \ref{tab:my-table3}, our framework demonstrates a significant improvement from the previous models in all three metrics when the level of missing data is smaller than 80\%. In particular, when the rate of missing data is 20\%, there is about a five-fold improvement in the accuracy. 
When 90\% of the original data is missing, our method shows a comparable performance to prior methods in all three metrics.


\subsection{Forecasting with Complete Data.} \label{subsec_forecast}

\begin{figure}[ht]
    \centering
    \begin{subfigure}{0.329\linewidth}
    \centering
    \includegraphics[width=\textwidth]{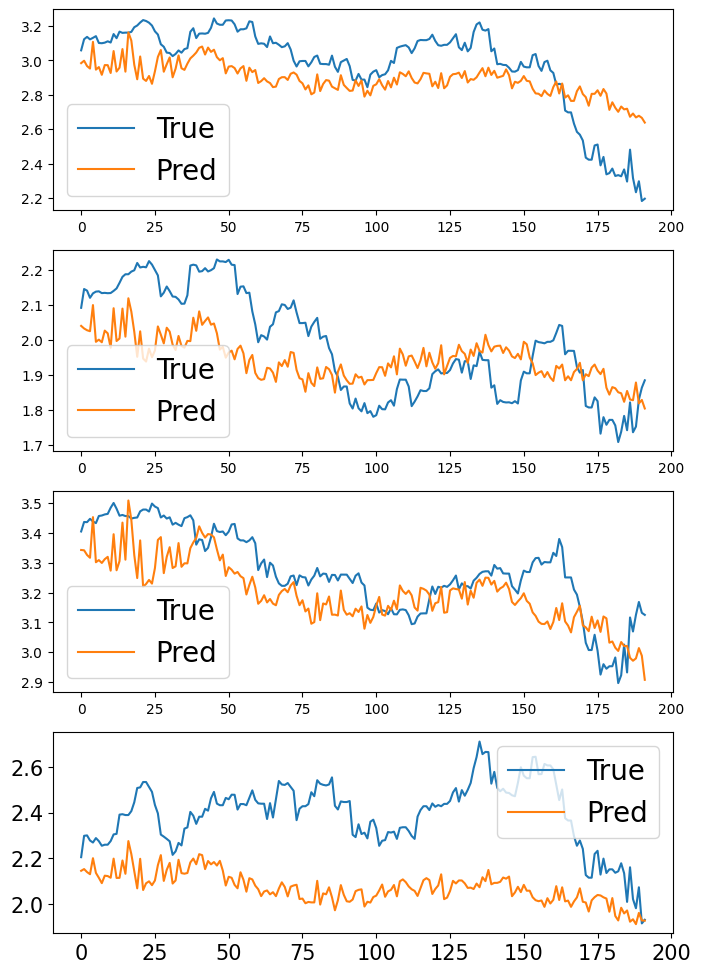}
    \caption{Exchange dataset}
    \end{subfigure}
    \begin{subfigure}{0.329\linewidth}
    \centering
    \includegraphics[width=\textwidth]{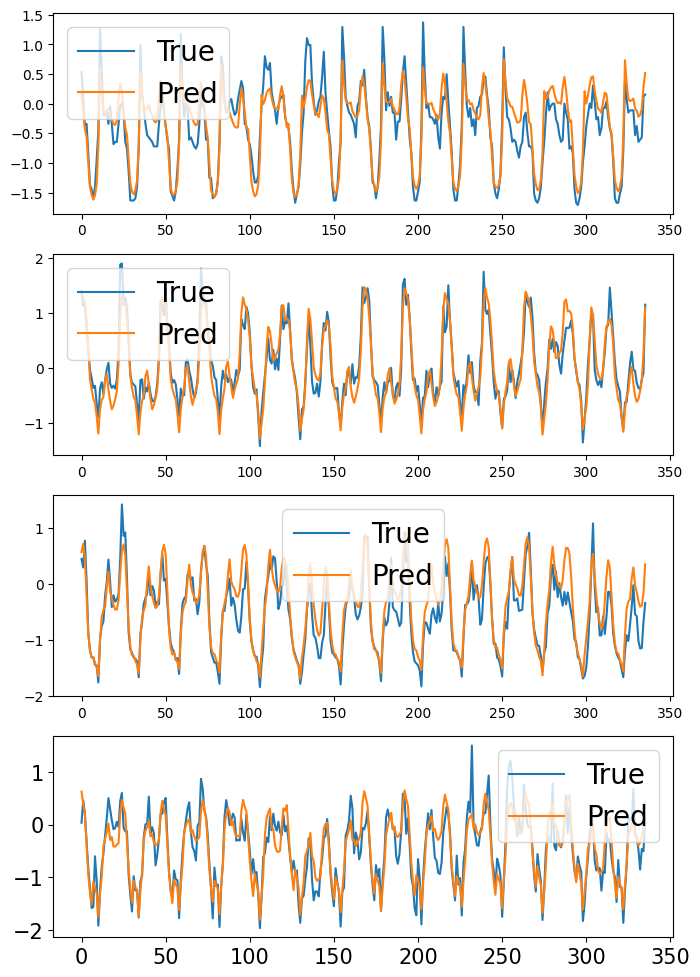}
    \caption{Electricity dataset}
    \end{subfigure}
    \begin{subfigure}{0.329\linewidth}
    \centering
    \includegraphics[width=\textwidth]{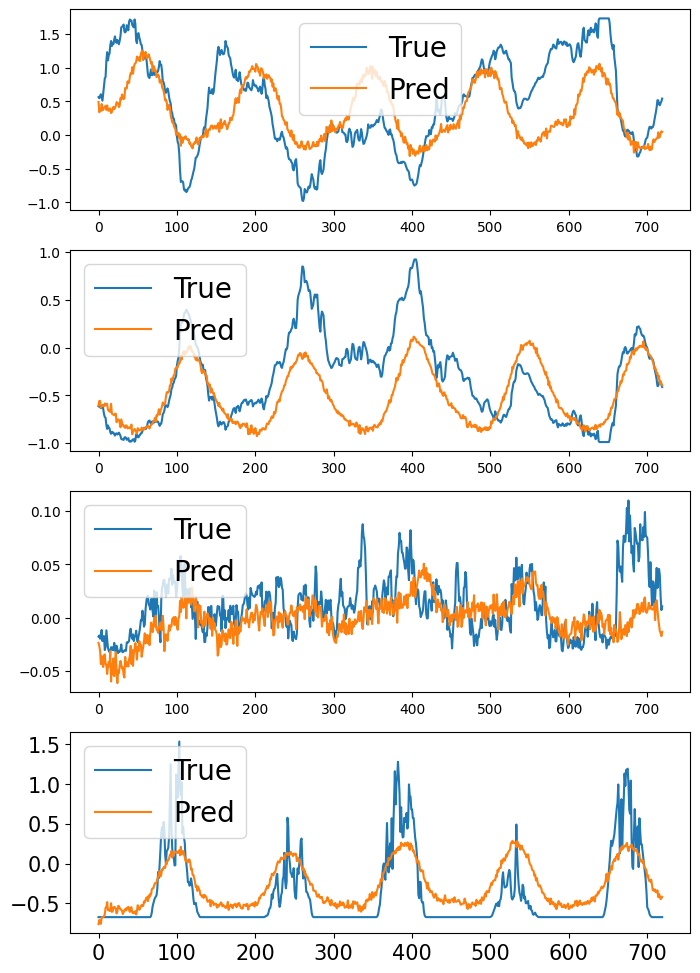}
    \caption{Weather dataset}
    \end{subfigure}
    \caption{Predictions from our model for Exchange dataset (in the left), Electricity dataset (in the middle) and Weather dataset (in the right). Each plot shows the prediction of one feature, and we show different output lengths for better demonstration of our results (over 192 time steps for Exchange data, 336 time steps for Electricity data and 720 time steps for Weather data). The Exchange dataset has 8 variables, while the Weather dataset has 21 features and the Electricity dataset has 321 variables.}
    \label{fig:plot_main}
\end{figure}


\textbf{Settings.} 
In this section, we present the forecasting experiments for five real-world datasets: Traffic, Electricity, Weather, ILI, Exchange-Rate  \citep{autoformer}. 
Electricity dataset records electricity consumption of 321 customers every hour in three years. Exchange dataset contains the exchange rates of eight different countries everyday from 1990 to 2016. Traffic dataset records hourly data from California Department of Transportation which contains the road occupancy rates measured by various sensors on San Francisco Bay area. 
Weather dataset contains 21 meteorological indicators and is recorded every 10 minutes in 2020. ILI dataset includes the influenza-like illness (ILI) patients data from Centers for Disease Control and Prevention of the United States every week between 2002 and 2021. 
We follow the standard experiment procedure as in \citep{autoformer}. The time series data is split into training, validation and test sets in chronological order by the ratio of 7:1:2 for all the data sets. 
We select approximately 14\% of the last features to be the features of interest $\bm{Y}$.
For example, since ILI dataset has 7 features, we let $d_x = 6$ and $d_y=1$ where the last feature represents the feature $\bm{Y}$. This choice aligns well with some of the univariate forecasting problems where the last features is often of interest \citep{autoformer}. The model is trained using PyTorch \citep{pytorch} with ADAM optimizer \citep{adam} for 10 training epochs, which is consistent with prior references \citep{informer,autoformer}.
We report the MAE and MSE on test data $\bm{Z}_O$, lower metrics indicate better results.
The experiments are repeated five times, and 
and we show the average results. 

\renewcommand{\arraystretch}{1.15}
\setlength{\tabcolsep}{2pt}
\begin{table}[ht]
\caption{
Comparison of forecasting performance for various methods on five real-world datasets. We highlight {\color{red}\textbf{the best}} results and {\color{blue}\underline{the second best}} results.
For the ILI dataset, $I = 36$ and $O \in \{24, 36, 48, 60\}$. For other datasets, $I = 96$ and $O \in \{96, 192, 336, 720\}$.
The results of other methods are from their reference, we only implement our method. We omit some prior works that perform worse due to space limit.}
\label{tab:my-table}
\begin{center}
\begin{adjustbox}{width=1\textwidth}
\begin{tabular}{cc *{7}{cc}}
\toprule[1pt]\midrule[0.3pt]
 &  & \multicolumn{2}{c}{LogTrans} & \multicolumn{2}{c}{Informer} & \multicolumn{2}{c}{Autoformer} & \multicolumn{2}{c}{FEDformer} & \multicolumn{2}{c}{DLinear-I} & \multicolumn{2}{c}{DLinear-S} & \multicolumn{2}{c}{Our method} \\
\multirow{-2}{*}{{Data}} & \multirow{-2}{*}{$O$} & MSE & MAE & MSE & MAE & MSE & MAE & MSE & MAE & MSE & MAE & MSE & MAE & MSE & MAE \\
\hline
 & 96 & 0.258 & 0.357 & 0.274 & 0.368 & 0.201 & 0.317 & 0.192 & 0.308 & {\color{blue}\underline{0.184}} & {\color{blue}\underline{0.270}} & 0.194 & 0.276 & {\color{red}\textbf{0.166}} & {\color{red}\textbf{0.257}} \\
 & 192 & 0.266 & 0.368 & 0.296 & 0.386 & 0.222 & 0.334 & 0.201 & 0.315 & {\color{blue}\underline{0.184}} & {\color{blue}\underline{0.273}} & 0.193 & 0.280 & {\color{red}\textbf{0.173}} & {\color{red}\textbf{0.265}} \\
 & 336 & 0.280 & 0.380 & 0.300 & 0.394 & 0.231 & 0.338 & 0.214 & 0.329 & {\color{blue}\underline{0.197}} & {\color{blue}\underline{0.289}} & 0.206 & 0.296 & {\color{red}\textbf{0.190}} & {\color{red}\textbf{0.283}} \\
\multirow{-4}{*}{{Electricity}} & 720 & 0.283 & 0.376 & 0.373 & 0.439 & 0.254 & 0.361 & 0.246 & 0.355 & {\color{blue}\underline{0.234}} & {\color{blue}\underline{0.323}} & 0.242 & 0.329 & {\color{red}\textbf{0.226}} & {\color{red}\textbf{0.316}} \\
\hline
 & 96 & 0.968 & 0.812 & 0.847 & 0.752 & 0.197 & 0.323 & 0.148 & 0.278 & 0.084 & 0.216 & {\color{red}\textbf{0.078}} & {\color{red}\textbf{0.197}} & {\color{blue}\underline{0.083}} & {\color{blue}\underline{0.206}} \\
 & 192 & 1.040 & 0.851 & 1.204 & 0.895 & 0.300 & 0.369 & 0.271 & 0.380 & {\color{blue}\underline{0.157}} & 0.298 & 0.159 & {\color{blue}\underline{0.292}} & {\color{red}\textbf{0.155}} & {\color{red}\textbf{0.288}} \\
 & 336 & 1.659 & 1.081 & 1.672 & 1.036 & 0.509 & 0.524 & 0.460 & 0.500 & {\color{red}\textbf{0.236}} & {\color{red}\textbf{0.379}} & 0.274 & 0.391 & {\color{blue}\underline{0.260}} & {\color{blue}\underline{0.383}} \\
\multirow{-4}{*}{{Exchange}} & 720 & 1.941 & 1.127 & 2.478 & 1.310 & 1.447 & 0.941 & 1.195 & 0.841 & {\color{blue}\underline{0.626}} & 0.634 & {\color{red}\textbf{0.558}} & {\color{red}\textbf{0.574}} & 0.661 & {\color{blue}\underline{0.619}} \\
\hline
 & 96 & 0.684 & 0.384 & 0.719 & 0.391 & 0.613 & 0.388 & {\color{blue}\underline{0.587}} & {\color{blue}\underline{0.366}} & 0.647 & 0.403 & 0.650 & 0.396 & {\color{red}\textbf{0.491}} & {\color{red}\textbf{0.308}} \\
 & 192 & 0.685 & 0.390 & 0.696 & 0.379 & 0.616 & 0.382 & 0.604 & 0.373 & 0.602 & 0.375 & {\color{blue}\underline{0.598}} & {\color{blue}\underline{0.370}} & {\color{red}\textbf{0.496}} & {\color{red}\textbf{0.312}} \\
 & 336 & 0.734 & 0.408 & 0.777 & 0.420 & 0.622 & {\color{blue}\underline{0.337}} & 0.621 & 0.383 & 0.607 & 0.377 & {\color{blue}\underline{0.605}} & 0.373 & {\color{red}\textbf{0.514}} & {\color{red}\textbf{0.318}} \\
\multirow{-4}{*}{{Traffic}} & 720 & 0.717 & 0.396 & 0.864 & 0.472 & 0.660 & 0.408 & {\color{blue}\underline{0.626}} & {\color{blue}\underline{0.382}} & 0.646 & 0.398 & 0.645 & 0.394 & {\color{red}\textbf{0.553}} & {\color{red}\textbf{0.337}} \\
\hline
 & 96 & 0.458 & 0.490 & 0.300 & 0.384 & 0.266 & 0.336 & 0.217 & 0.296 & {\color{red}\textbf{0.164}} & {\color{blue}\underline{0.237}} & 0.196 & 0.255 & {\color{blue}\underline{0.168}} & {\color{red}\textbf{0.227}} \\
 & 192 & 0.658 & 0.589 & 0.598 & 0.544 & 0.307 & 0.367 & 0.276 & 0.336 & {\color{red}\textbf{0.209}} & {\color{blue}\underline{0.282}} & 0.237 & 0.296 & {\color{blue}\underline{0.211}} & {\color{red}\textbf{0.269}} \\
 & 336 & 0.797 & 0.652 & 0.578 & 0.523 & 0.359 & 0.395 & 0.339 & 0.380 & {\color{blue}\underline{0.263}} & {\color{blue}\underline{0.327}} & 0.283 & 0.335 & {\color{red}\textbf{0.261}} & {\color{red}\textbf{0.309}} \\
\multirow{-4}{*}{{Weather}} & 720 & 0.869 & 0.675 & 1.059 & 0.741 & 0.419 & 0.428 & 0.403 & 0.428 & {\color{blue}\underline{0.338}} & {\color{blue}\underline{0.380}} & 0.345 & 0.381 & {\color{red}\textbf{0.332}} & {\color{red}\textbf{0.363}} \\
\hline
 & 24 & 4.480 & 1.444 & 5.764 & 1.677 & 3.483 & 1.287 & 3.228 & 1.260 & 3.015 & 1.192 & {\color{red}\textbf{2.398}} & {\color{red}\textbf{1.040}} & {\color{blue}\underline{2.657}} & {\color{blue}\underline{1.142}} \\
 & 36 & 4.799 & 1.467 & 4.755 & 1.467 & 3.103 & 1.148 & 2.679 & 1.080 & 2.737 & {\color{red}\textbf{1.036}} & {\color{blue}\underline{2.646}} & 1.088 & {\color{red}\textbf{2.326}} & {\color{blue}\underline{1.046}} \\
 & 48 & 4.800 & 1.468 & 4.763 & 1.469 & 2.669 & 1.085 & 2.622 & 1.078 & {\color{blue}\underline{2.577}} & {\color{red}\textbf{1.043}} & 2.614 & 1.086 & {\color{red}\textbf{2.469}} & {\color{blue}\underline{1.070}} \\
\multirow{-4}{*}{{ILI}} & 60 & 5.278 & 1.560 & 5.264 & 1.564 & {\color{blue}\underline{2.770}} & 1.125 & 2.857 & 1.157 & 2.821 & {\color{red}\textbf{1.091}} & 2.804 & 1.146 & {\color{red}\textbf{2.677}} & {\color{blue}\underline{1.119}}\\
\midrule[0.3pt]\bottomrule[1pt]
\end{tabular}
\end{adjustbox}
\end{center}
\end{table}

\textbf{Results.} Our model is compared with the following methods: DLinear-I and DLinear-S \citep{linearmodels}, FEDformer \citep{fedformer},
Autoformer \citep{autoformer}, Informer \citep{informer}, Pyraformer \citep{pyraformer}, LogTrans \citep{logtrans}. 
We compare with these benchmarks as they use the same experiment procedure in \citep{informer} with 10 training epochs. 
The results are shown in  Table \ref{tab:my-table}. 
This experiment demonstrates that our model outperforms previous models on Electricity and Traffic data, where the number of features is large. On the other datasets, our method still performs comparable to or slightly better than previous benchmark. We show our predictions in Figure \ref{fig:plot_main}. Although the Exchange and Weather datasets are difficult to predict, our long-term predictions still follow the overal trend of the test data.



\subsection{Forecasting With Missing Data}\label{subsec_forecast2}

From Table \ref{tab:my-table}, it can be seen that our method performs relatively good for three datasets Traffic, Electricity and Weather. 
Now, we investigate the model performance for a forecasting task with missing data.
We proceed to conduct a forecasting experiment on these three datasets but with missing values. We randomly remove 10 to 40 percent of the data $\bm{Z}$ of Traffic, Electricity, and Weather datasets for the purpose of this experiment. 
Note that while the training model only have access to some level of data $\bm{Z}^*$, the test model compute the metrics based on complete ground truth, which indicates fair comparisons.

\begin{table}[ht]
\caption{Forecasting performance with various level of missing rates, and comparison with the second-best results. We report {\color{blue}\underline{the second best}} results in Table \ref{tab:my-table} in the last column. 
We highlight the results returned by missing data which perform {\color{blue}\textbf{better than or similar to}} the second-best results in each row.  }
\begin{center}
\label{tab:my-table2}
\begin{tabular}{cccccccccccccc}
\toprule[1pt]\midrule[0.3pt]
 &  & \multicolumn{2}{c}{0\%} & \multicolumn{2}{c}{10\%} & \multicolumn{2}{c}{20\%} & \multicolumn{2}{c}{30\%} & \multicolumn{2}{c}{40\%} & \multicolumn{2}{c}{Second} \\
\multirow{-2}{*}{{Data}} & \multirow{-2}{*}{$O$} & MSE & MAE & MSE & MAE & MSE & MAE & MSE & MAE & MSE & MAE & MSE & MAE \\
\hline
 & 96 & {\color{red}\textbf{0.166}} & {\color{red}\textbf{0.257}} & {\color{blue}\textbf{0.176}} & {\color{blue}\textbf{0.268}} & {\color{blue}\textbf{0.183}} & 0.276 & 0.189 & 0.283 & 0.194 & 0.288 & {\color{blue}\underline{0.184}} & {\color{blue}\underline{0.270}} \\
 & 192 & {\color{red}\textbf{0.173}} & {\color{red}\textbf{0.265}} & {\color{blue}\textbf{0.183}} & 0.277 & 0.190 & 0.285 & 0.196 & 0.291 & 0.201 & 0.296 & {\color{blue}\underline{0.184}} & {\color{blue}\underline{0.273}} \\
 & 336 & {\color{red}\textbf{0.190}} & {\color{red}\textbf{0.283}} & 0.199 & 0.294 & 0.206 & 0.301 & 0.211 & 0.306 & 0.216 & 0.311 & {\color{blue}\underline{0.197}} & {\color{blue}\underline{0.289}} \\
\multirow{-4}{*}{{Electricity}} & 720 & {\color{red}\textbf{0.226}} & {\color{red}\textbf{0.316}} & 0.235 & 0.326 & 0.240 & 0.331 & 0.246 & 0.337 & 0.250 & 0.340 & {\color{blue}\underline{0.234}} & {\color{blue}\underline{0.323}} \\
\hline
 & 96 & {\color{red}\textbf{0.491}} & {\color{red}\textbf{0.308}} & {\color{blue}\textbf{0.522}} & {\color{blue}\textbf{0.325}} & {\color{blue}\textbf{0.545}} & {\color{blue}\textbf{0.335}} & {\color{blue}\textbf{0.563}} & {\color{blue}\textbf{0.344}} & {\color{blue}\textbf{0.580}} & {\color{blue}\textbf{0.352}} & {\color{blue}\underline{0.587}} & {\color{blue}\underline{0.366}} \\
 & 192 & {\color{red}\textbf{0.496}} & {\color{red}\textbf{0.312}} & {\color{blue}\textbf{0.525}} & {\color{blue}\textbf{0.329}} & {\color{blue}\textbf{0.546}} & {\color{blue}\textbf{0.336}} & {\color{blue}\textbf{0.564}} & {\color{blue}\textbf{0.341}} & {\color{blue}\textbf{0.578}} & {\color{blue}\textbf{0.347}} & {\color{blue}\underline{0.598}} & {\color{blue}\underline{0.370}} \\
 & 336 & {\color{red}\textbf{0.514}} & {\color{red}\textbf{0.318}} & {\color{blue}\textbf{0.538}} & {\color{blue}\textbf{0.335}} & {\color{blue}\textbf{0.561}} & 0.340 & {\color{blue}\textbf{0.577}} & 0.348 & {\color{blue}\textbf{0.591}} & 0.351 & {\color{blue}\underline{0.605}} & {\color{blue}\underline{0.337}} \\
\multirow{-4}{*}{{Traffic}} & 720 & {\color{red}\textbf{0.553}} & {\color{red}\textbf{0.337}} & {\color{blue}\textbf{0.583}} & {\color{blue}\textbf{0.348}} & {\color{blue}\textbf{0.605}} & {\color{blue}\textbf{0.356}} & {\color{blue}\textbf{0.621}} & {\color{blue}\textbf{0.362}} & 0.636 & {\color{blue}\textbf{0.368}} & {\color{blue}\underline{0.626}} & {\color{blue}\underline{0.382}} \\
\hline
 & 96 & {\color{blue}\underline{0.168}} & {\color{red}\textbf{0.227}} & 0.178 & 0.239 & 0.182 & 0.246 & 0.184 & 0.249 & 0.185 & 0.249 & {\color{blue}\underline{0.168}} & {\color{blue}\underline{0.237}} \\
 & 192 & {\color{blue}\underline{0.211}} & {\color{red}\textbf{0.269}} & 0.220 & {\color{blue}\textbf{0.279}} & 0.225 & 0.286 & 0.228 & 0.288 & 0.228 & 0.288 & {\color{blue}\underline{0.211}} & {\color{blue}\underline{0.282}} \\
 & 336 & {\color{red}\textbf{0.261}} & {\color{red}\textbf{0.309}} & 0.269 & {\color{blue}\textbf{0.318}} & 0.277 & 0.330 & 0.277 & {\color{blue}\textbf{0.326}} & 0.280 & 0.330 & {\color{blue}\underline{0.263}} & {\color{blue}\underline{0.327}} \\
\multirow{-4}{*}{{Weather}} & 720 & {\color{red}\textbf{0.332}} & {\color{red}\textbf{0.363}} & {\color{blue}\textbf{0.338}} & {\color{blue}\textbf{0.369}} & 0.340 & {\color{blue}\textbf{0.371}} & 0.343 & {\color{blue}\textbf{0.374}} & 0.345 & {\color{blue}\textbf{0.376}} & {\color{blue}\underline{0.338}} & {\color{blue}\underline{0.380}}

\\
\midrule[0.3pt]\bottomrule[1pt]
\end{tabular}
\end{center}
\end{table}

\textbf{Results.} 
Table \ref{tab:my-table2} shows the forecasting performance of our model with 10 $\sim$ 40 percent of missing data. For a comparison, we show the forecasting errors of the second best models in Table \ref{tab:my-table}. The forecasting accuracy decreases as the fraction of missing entries increases, which matches perfectly with our intuition. However, our model performs reasonably well comparing with other methods, even with the missing entries. For example, for Traffic data, our model makes a better prediction even with 40\% missing values than the Transformer models, such as \citep{fedformer, autoformer, informer}, with the complete data, \emph{i.e.}, without a missing entry.


\section{Model Analysis and Ablation Study}\label{sec_ablation}
In this section, we consider the forecasting experiment with complete data as in Section \ref{subsec_forecast}. In order to demonstrate the efficiency of our learner functions $f$ and $g$ with our loss function, we compare our approach with other modeling options. Table \ref{tab:my-table4} shows our model analysis, and ablation study of the components on the loss function. 
In the first experiment, we keep all the training details and trained model unchanged, but perform the test without the information of data $\bm{Y}$. The model predicts the data $\bm{Y}$ using the inference function $f$ and then use $\{\bm{X}, f(\bm{X})\}$ for forecasting task. 

In a modelling perspective, 
we note that the approach in Section \ref{subsec_learner} use the spatial relation learner function $\hat{f}$ across all the time steps, \textit{i.e.} $\hat{f}(\bm{x}^*_i) = \bm{y_i}$. Similarly, the temporal relation function $\hat{g}$ that takes an $I$-dimensional vector as an input and generates a $(I+O)$-dimensional vector for an output is used across all the features. Then we parameterize $\hat{f}$ and $\hat{g}$ by fully connected networks (FCNs). 
Thus, instead of restricting the model structures as $\hat{f}$ or $\hat{g}$, we use FCNs that takes the flattened arrays of $\bm{X}^*$ and $\bm{Z}^*$ as inputs and generates $\bm{Y}$ and $\bm{Z}_+$. We refer to this approach as ``No broadcast'' in Table \ref{tab:my-table4}. Next, to analyze the choice of the fully connected networks, we replace the FCNs of $\hat{g}$ and $\hat{f}$ by linear functions and keep other factors unchanged. This experiment is called the ``No FCNs'' approach in Table \ref{tab:my-table4}. 

In addition, we compare our loss function with the individual training of forecasting task only. In the third experiment ``No $Y= f(X)$" we study the ablation where the training loss does not contain the component that minimizes the distance of $\bm{Y}$ and $f(\bm{X}^*)$, and other implementation details are unchanged. Similarly, in the option ``No imputation" we experiment with the loss function that only contains the forecast element \textit{i.e.} 
\begin{align*}
    L (\bm{Z}) &= \|(\bm{Y}^* - f (\bm{X}^*)) \odot \bm{M}_{Y^*}\|_F^2 +  \|(\bm{Z}_{O}^* - g' (\{\bm{X}^*, f(\bm{X}^*)\}) \odot \bm{M}_{Z_{O}^*}\|_F^2 + r(w_f,w_g), 
\end{align*}
where $\bm{Z} = \{\bm{X}, \bm{Y}\}$ and $g'$ is a corresponding learner function that maps from $ \mathbb{R}^{I\times d}$ to $\mathbb{R}^{O\times d}$ (instead of $g$ that maps from $ \mathbb{R}^{I\times d}$ to $\mathbb{R}^{(I+O)\times d}$). We keep other implementation details the same. 
Due to the space limitation, we only report the results of Electricity, Weather and Traffic data in Table \ref{tab:my-table4}. The results for other data sets are similar. 

\begin{table}[ht]
\caption{
Analysis of different choices for the dynamic model, and   
ablation of the components on the loss function. We highlight {\color{red}\textbf{the best}} results and {\color{blue}\underline{the second best}} results.
}
\begin{center}
\label{tab:my-table4}
\begin{tabular}{cccccccccccccccc}
\toprule[1pt]\midrule[0.3pt]
 &  & \multicolumn{2}{c}{Our method} & \multicolumn{2}{c}{No $Y$ data} & \multicolumn{2}{c}{No broadcast} & \multicolumn{2}{c}{No FCNs} & \multicolumn{2}{c}{No $Y= f(X)$} & \multicolumn{2}{c}{No imputation} \\
\multirow{-2}{*}{Data} & \multirow{-2}{*}{$O$} & MSE & MAE & MSE & MAE & MSE & MAE & MSE & MAE & MSE & MAE & MSE & MAE \\
\hline
 & 96 & {\color{red} \textbf{0.166}} & {\color{red} \textbf{0.257}} & {\color{blue}\underline{ 0.171}} & {\color{blue}\underline{ 0.262}} & 0.333 & 0.409 & 0.195 & 0.278 & 0.179 & 0.266 & 0.174 & 0.264 \\
 & 192 & {\color{red} \textbf{0.173}} & {\color{red} \textbf{0.265}} & {\color{blue}\underline{ 0.179}} & {\color{blue}\underline{ 0.270}} & 0.339 & 0.417 & 0.194 & 0.281 & 0.191 & 0.281 & 0.181 & 0.273 \\
 & 336 & {\color{red} \textbf{0.190}} & {\color{red} \textbf{0.283}} & {\color{blue}\underline{ 0.194}} & {\color{blue}\underline{ 0.289}} & 0.350 & 0.427 & 0.207 & 0.296 & 0.202 & 0.293 & 0.195 & {\color{blue}\underline{ 0.289}} \\
\multirow{-4}{*}{\rotatebox{90}{Electricity}} & 720 & {\color{red} \textbf{0.226}} & {\color{red} \textbf{0.316}} & {\color{blue}\underline{ 0.232}} & 0.324 & 0.357 & 0.432 & 0.242 & 0.329 & 0.236 & {\color{blue}\underline{ 0.322}} & 0.234 & 0.324 \\
\hline
 & 96 & {\color{red} \textbf{0.168}} & {\color{red} \textbf{0.227}} & {\color{blue}\underline{ 0.173}} & 0.231 & 0.185 & 0.276 & 0.200 & 0.261 & 0.174 & {\color{blue}\underline{ 0.230}} & 0.175 & 0.245 \\
 & 192 & {\color{red} \textbf{0.211}} & {\color{red} \textbf{0.269}} & 0.213 & {\color{blue}\underline{ 0.270}} & 0.236 & 0.319 & 0.241 & 0.299 & 0.216 & 0.273 & {\color{red} \textbf{0.211}} & {\color{blue}\underline{ 0.270}} \\
 & 336 & {\color{red} \textbf{0.261}} & {\color{red} \textbf{0.309}} & {\color{blue}\underline{ 0.265}} & 0.313 & 0.311 & 0.372 & 0.287 & 0.336 & 0.266 & {\color{blue}\underline{ 0.310}} & {\color{blue}\underline{ 0.265}} & 0.315 \\
\multirow{-4}{*}{\rotatebox{90}{Weather}} & 720 & {\color{red} \textbf{0.332}} & {\color{red} \textbf{0.363}} & {\color{blue}\underline{ 0.336}} & 0.367 & 0.417 & 0.445 & 0.351 & 0.387 & 0.338 & {\color{blue}\underline{ 0.366}} & 0.338 & 0.371 \\
\hline
 & 96 & {\color{red} \textbf{0.491}} & {\color{red} \textbf{0.308}} & {\color{blue}\underline{ 0.512}} & {\color{blue}\underline{ 0.338}} & 0.893 & 0.786 & 0.652 & 0.398 & 0.520 & 0.382 & 0.523 & 0.359 \\
 & 192 & {\color{red} \textbf{0.496}} & {\color{red} \textbf{0.312}} & {\color{blue}\underline{ 0.533}} & {\color{blue}\underline{ 0.346}} & 0.994 & 0.713 & 0.600 & 0.371 & 0.548 & 0.352 & 0.539 & 0.369 \\
 & 336 & {\color{red} \textbf{0.514}} & {\color{red} \textbf{0.318}} & {\color{blue}\underline{ 0.542}} & {\color{blue}\underline{ 0.367}} & 1.021 & 0.845 & 0.606 & 0.374 & 0.557 & 0.373 & 0.548 & 0.371 \\
\multirow{-4}{*}{\rotatebox{90}{Traffic}} & 720 & {\color{red} \textbf{0.553}} & {\color{red} \textbf{0.337}} & 0.562 & 0.383 & 1.341 & 0.967 & 0.646 & 0.396 & {\color{blue}\underline{ 0.561}} & 0.389 & 0.569 & {\color{blue}\underline{ 0.381}}\\
\midrule[0.3pt]\bottomrule[1pt]
\end{tabular}
\end{center}
\end{table}

\textbf{Results.} In the first experiment, the test model does not have the complete information of $\bm{Y}$, thus intuitively it may perform worse than forecasting directly from the true data. While the results reflect this point, they also suggest that the trained model predict reasonably well even when the data $\bm{Y}$ is not available \textit{i.e.} the inference model learns the target $\bm{Y}$ effectively. The "No broadcast" experiments  confirm that restricting the structures of the spatial learner and temporal learner functions performs much better than the other flattening approach, which contributes significantly to the performance of our method.
The next experiments "No FCNs" show that using FCNs for $\hat{f}$ and $\hat{g}$ demonstrates better performance than using linear models. Moreover, the last two ablation studies demonstrate that our loss function which learns the inference relation $\bm{Y}= f(\bm{X}^*)$ and the imputation task simultaneously is indeed beneficial to the training progress of the forecasting task.

\section{Conclusions.}
We propose an end-to-end multi-task framework for time series data, which simultaneously imputes the missing entries and makes a multiple-step ahead prediction. 
Our framework is inspired by the idea of 
learning the mapping from auxiliary features to target features in the presence of missing entries, with the goal to leverage this information to predict future system states. This capability is particularly valuable as it enables prediction even with observations containing missing values.
Through extensive experiments, our framework demonstrates strong performance in both imputation and forecasting tasks. 
While our framework offers flexibility in employing various architectures, we propose a specific structure for the spatial learner and temporal learner functions to regulate the behavior of the model. Our ablation study confirms the effectiveness of training with combined losses exceeding the performance achieved by individual losses.





\appendix

\appendix
\newpage

\section*{Appendix}

\section{Dealing with missing data}

Our framework is flexible that it can handle two cases whether the data has missing values or not. 
Without loss of generality, let us assume that the original missing rate is $r_{\text{orig}}$. If the data has no missing entries, then $r_{\text{orig}} = 0$. The model processes every data matrix with an additional mask that indicate which value is available. 
There are several reasons why we choose to mask the data artificially: 
\begin{itemize}
    \item Firstly (in validation and test set), to test the performance of the model. For this purpose, we use the artificially masked data to make prediction, then use the all available data as ground truth. We use artificial rate  $r_{\text{arti}}$ here. 
    \item Secondly (in train set), we add artificial mask
    to test the model for different levels of missing data. This artificially masked data is the only information that is available in the training progress. We also use the artificial rate  $r_{\text{arti}}$ here. 
    \item Finally (in training progress), we may choose to mask some values and use these values as ground truth/ incentive for the model to learn the correct data. For this purpose, the input has masked values, however the loss function use training data to make predictions. Since this option is related to the model, we may use a different rate independent of the problem (which is a hyper parameter chosen by the training process). This rate is denoted as $r_{\text{input mask}}$.    
\end{itemize}
\section{Imputation Task Implementation}
The implementation of our imputation experiment follows the prior work \cite{saits} where the repository is published at \url{https://github.com/WenjieDu/SAITS}.

\textbf{Datasets}.

The Electricity Load Diagrams dataset is a public dataset from UCI that contains electricity consumption data collected from 370 clients every 15 minutes. The data covers the period from 2011/01/01 to 2014/12/31. The dataset has no missing data, \textit{i.e.} $r_{\text{orig}} = 0$.
To process the dataset, we split it into three sets: a training set, a validation set, and a test set. The training set consists of the data from 2012/09 to 2014/12. The validation set consists of the data from 2011/11 to 2012/08. The test set consists of the data from 2011/01 to 2011/10.
To generate time-series data for model training, we select every 100 consecutive steps from the training set. The data has a total of 1400 samples and 370 features.
Since the dataset has no missing values, we artificially add missing values to the training set, validation set, and test set. The missing rate varies from 10\% to 90\%, \textit{i.e.} $r_{\text{arti}}$ takes values from 0.1 to 0.9. Results for 20\% to 90\% are in Table \ref{tab:my-table3} while results for 10\% are in Table \ref{tab:my-table3.1}. 

\textbf{Evaluation metrics.} 

We use three metrics to evaluate the imputation performance: Mean Absolute Error (MAE), Root Mean Square Error(RMSE), and Mean Relative Error (MRE). 
Let $P \in \R^{D \times I}$ be the predicted value of our model and $V\in \R^{D \times I}$ be the ground truth value that may contain missing values. The masked matrix $M$ indicates whether the entries of $V$ are missing or not. We only compute the errors with the masked matrix $M$ as there is no information of the missing entries. The metrics are presented as follows: 
\begin{align*}
\text{MAE}(P, V) & =\frac{\sum_{d=1}^D \sum_{t=1}^I | P_t^d-   V_t^d| \times   M_t^d}{\sum_{d=1}^D \sum_{t=1}^I   M_t^d}, \\
\text{RMSE}(P, V) & =\sqrt{\frac{\sum_{d=1}^D \sum_{t=1}^I( P_t^d-   V_t^d)^2 \times   M_t^d}{\sum_{d=1}^D \sum_{t=1}^I   M_t^d}}, \\
\text{MRE}(P, V) & =\frac{\sum_{d=1}^D \sum_{t=1}^I | P_t^d-   V_t^d| \times   M_t^d}{\sum_{d=1}^D \sum_{t=1}^I   | V_t^d| \times  M_t^d}.
\end{align*}

\textbf{Training details.}

We follow the implementation of SAITS that set the batch size to 128. The number of training epochs is 30. In Table \ref{tab:my-table3} and Table \ref{tab:my-table3.1}, the results for two na\"{i}ve imputation method (Last and Median), and five state-of-the-art models 
M-RNN \citep{mrnn}; GP-VAE \citep{gpvae}; BRITS \citep{brits}, Transformer \citep{attention} and SAITS \citep{saits} are from the reference \citep{saits}. Those methods use early stopping strategy when the MAE does not decrease. 
In the meantime, we report the model with the highest validation performance. For our imputation training, we set $r_{\text{input mask}}$ to be 0.5 as withholding some of the values help the model learn the imputation task faster. 
We use L2 regularized loss for all of our experiments. In addition, we impose another regularizer to help smoothing out the prediction. 
We also report the 
model parameters of the benchmarks in Table \ref{tab:my-table3.1}. Our number of parameter is moderate but we still obtain better results than prior works. We do not report training time as we use different computational resources compared to prior benchmarks. 

\begin{table}[ht]
\caption{Comparison of the imputation accuracy for the Electricity data, and report of the model parameters of the benchmarks. The level of the missing data in this table is 10\%. Results for 20\% to 90\% are in Table \ref{tab:my-table3}. We report the mean absolute error (MAE), root mean square error (RMSE), and mean relative error (MRE), where lower metrics demonstrate better results.
We highlight {\color{red}\textbf{the best}} results and {\color{blue}\underline{the second best}} results.
}
\label{tab:my-table3.1}
\begin{center}
\renewcommand{\arraystretch}{1.15}
\begin{tabular}{ccccc}
\toprule[1pt]\midrule[0.3pt]
\textbf{Measures} & \textbf{MAE} & \textbf{RMSE} & \textbf{MRE} & \textbf{Number of parameters} \\
\hline
{\color{red} \textbf{Our method}} & {\color{red} \textbf{0.192}} & {\color{red} \textbf{0.255}} & {\color{red} \textbf{10.7\%}} & 2.28M \\
SAITS & {\color{blue} \underline{ 0.735}} & {\color{blue} \underline{ 1.162}} & {\color{blue} \underline{ 39.4\%}} & 5.32M  \\
Transformer & 0.823 & {  1.301} & {  44.0\%} & 4.36M  \\
BRITS & 0.847 & {  1.322} & {  45.3\%} & 0.73M  \\
GP-VAE & 1.094 & {  1.565} & {  58.6\%} & 0.15M  \\
M-RNN & 1.244 & {  1.867} & {  66.6\%} & 0.07M  \\
Last & 1.006 & {  1.533} & {  53.9\%} & -  \\
Median & 2.056 & {  2.732} & {  110.1\%} & - \\
\midrule[0.3pt]\bottomrule[1pt]
\end{tabular}
\end{center}
\end{table}

\begin{table}[ht]
\caption{Description of the imputation model with three linear layers for Electricity imputation dataset. }
\label{tab:my-table3.2}
\renewcommand{\arraystretch}{1.15}
\begin{center}
\begin{tabular}{lm{6em}m{6em}m{6em}m{6em}}
\toprule[1pt]\midrule[0.3pt]
{ \textbf{Spatial learner $\hat{f}$}} & { \textbf{Input}} & { \textbf{Output}} & \multicolumn{2}{l}{{ \textbf{Parameter ranges}}} \\
\hline
Linear layer 1 & { 320} & { 960} & { 320} & 320-1200 \\
ReLU layer \\
Linear layer 2 & 960 & { 800} & { 320-1200} & 320-1200 \\
ReLU layer \\
Linear layer 3 & 800 & { 50} & { 320-1200} & 50 \\
\hline\hline
\textbf{Temporal learner $\hat{g}$} & \textbf{Input} & { \textbf{Output}} & \multicolumn{2}{l}{{ \textbf{Parameter ranges}}} \\
\hline
Linear layer 1 & 100 & { 1110} & { 100} & 370-1200 \\
ReLU layer \\
Linear layer 2 & 1110 & { 800} & { 370-1200} & 370-1200 \\
ReLU layer \\
Linear layer 3 & 800 & { 200} & { 370-1200} & 200 \\
\midrule[0.3pt]\bottomrule[1pt]
\end{tabular}
\end{center}
\end{table}
 As presented in Section \ref{subsec_learner}, the inference function is invariant in time dimension and all the time step vectors share a common function $\hat{f}$. Similarly $g$ broadcasts a function $\hat{g}$ from dimension $I$ to dimension $I + O$. For the imputation datasets, we implement fully connected neural networks for $\hat{f}$ and $\hat{g}$ with three linear layers along with regular ReLU layers. Table \ref{tab:my-table3.2} describes our detailed architecture of the two learner functions used in our experiments. We search for the network configurations using a parameter range (in the last two columns), then we choose a final configuration for testing the model (reported in the first two columns). The parameter ranges are chosen as multiples of the number of total features (370) and the number of auxiliary variables $x$ (320), however, sometimes we cap the numbers at 1200 and 800 to prevent too large models. 

\begin{figure}[ht]
    \centering
    \begin{subfigure}{0.49\linewidth}
    \centering
    \includegraphics[width=\textwidth]{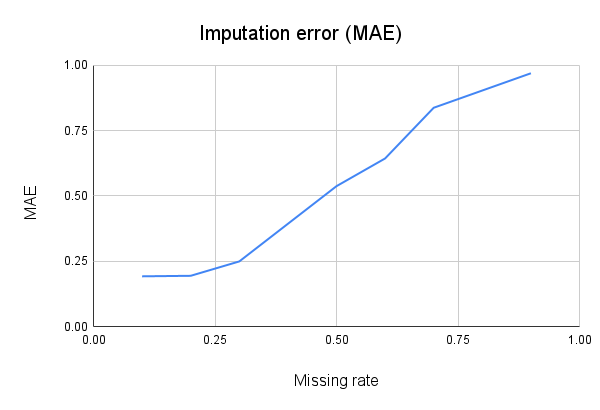}
    \caption{Imputation performance in MAE}
    \end{subfigure}
    \begin{subfigure}{0.49\linewidth}
    \centering
    \includegraphics[width=\textwidth]{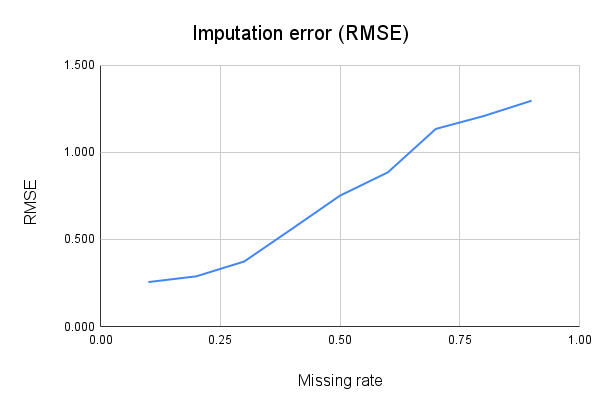}
    \caption{Imputation performance in RMSE}
    \end{subfigure}
    \caption{Imputation errors of our Framework when the missing rate varies from 10\% to 90\%.
    }
    \label{fig:impute}
\end{figure}

\textbf{Results analysis.}
Figure \ref{fig:impute} shows the imputation accuracy of our Framework when the missing rate varies from 10\% to 90\%. When the artificial missing rate increases linearly, the prediction error also increases almost linearly. 

\section{Forecasting Task Implementation}

The implementation of our imputation experiment follows the prior work \cite{autoformer} where the repository is published at \url{https://github.com/thuml/Autoformer}.

\textbf{Datasets}.
The original forecasting datasets have no missing data, \textit{i.e.} $r_{\text{orig}} = 0$. Below is the detailed description of each datasets. 
\begin{itemize}
    \item \textbf{Electricity Load Diagrams 2011 2014} can be found at \url{https://archive.ics.uci.edu/ml/datasets/ElectricityLoadDiagrams20112014}. This data has 321 features with 26304 time steps, which contains electric power consumption measurements for a group of clients. Each column represents one client, and each row represents one 15-minute period.The values are in kW. 
    On March 25th, there are only 23 hours in the day, and the values between 1:00 AM and 2:00 AM are zero for all clients. On October 31st, there are 25 hours in the day, and the values between 1:00 AM and 2:00 AM aggregate the consumption of two hours. In addition, some clients were created after 2011. For these clients, the consumption is considered to be zero.
    \item \textbf{Exchange Data} from the reference \cite{exchange} is the collection of the daily exchange rates of eight foreign countries including Australia, British, Canada, Switzerland, China, Japan, New Zealand and Singapore ranging from 1990 to 2016. This data has 8 features and 7588 time steps. 
    \item \textbf{California Traffic Data} was published by the Caltrans Performance Measurement System (PeMS) in the California Department of Transportation \url{https://pems.dot.ca.gov/}. This data has 862 features with 17544 time steps. It is collected in real-time from thousands of individual detectors spans the freeway system across all major metropolitan areas of the State of California.
    \item \textbf{Weather Data} can be found at \url{https://www.bgc-jena.mpg.de/wetter/}. This data has 21 features and 52696 time steps. Meteorological data was recorded every 10 minutes for the entire year 2020. The data contains 21 meteorological indicators, such as air temperature, pressure, humidity, radiation, and precipitation.
    \item \textbf{ILI - Outpatient Influenza-like Illness Data} was published at \url{https://gis.cdc.gov/grasp/fluview/fluportaldashboard.html}. This data has 7 features with 966 time steps. The Centers for Disease Control and Prevention (CDC) of the United States  collected weekly data on the number of patients seen with influenza-like illness (ILI) between 2002 and 2021. The data shows the ratio of patients seen with ILI to the total number of patients.
    
\end{itemize}
In the first forecasting experiment (Section \ref{subsec_forecast}), we train the model with the complete data where $r_{\text{orig}}= r_{\text{arti}} = 0$. In the next Section \ref{subsec_forecast2} we artificially add missing values to the training set, validation set, and test set where the missing rate varies from 10\% to 40\%, \textit{i.e.} $r_{\text{arti}}$ takes values from 0.1 to 0.4. 

\textbf{Evaluation metrics.} 

We report the MAE and MSE on test data $\bm{Z}_O$, lower metrics indicate better results.
The experiments are repeated five times, and and we show the average results. 
Similar to the imputation evaluation, let $P \in \R^{D \times O}$ be the predicted value of our model and $V\in \R^{D \times O}$ be the ground truth value. However, for the two forecasting tasks, we compute the errors using the complete data. This is different from the imputation task, and this helps to ensure fair comparison to prior works in Section \ref{subsec_forecast2} as all the experiments tested on complete ground truth data without missing values. The metrics are presented as follows: 
\begin{align*}
\text{MAE}(P, V) & = \frac{1}{D O}\sum_{d=1}^D \sum_{t=1}^O | P_t^d-   V_t^d|, \\
\text{MSE}(P, V) & =\frac{1}{D O}\sum_{d=1}^D \sum_{t=1}^O ( P_t^d-   V_t^d)^2.
\end{align*}
\textbf{Training details.}

We follow the implementation of Autoformer that set the batch size to 32. The number of training epochs is 10. In Table \ref{tab:my-table}the results of other methods DLinear-I and DLinear-S \citep{linearmodels}, FEDformer \citep{fedformer},
Autoformer \citep{autoformer}, Informer \citep{informer}, Pyraformer \citep{pyraformer}, LogTrans \citep{logtrans} are from their reference, we only implement our method.
We report the model with the highest validation performance within 10 training epochs.

Table \ref{tab:my-table.2} describes our detailed architecture of the two learner functions used in our forecasting experiments for Electricity, Traffic, Weather and 
ILI datasets. For Exchange dataset, since the linear functions $\hat{f}$ and $\hat{g}$ perform slightly better than the three-layer models, we do not choose multiple layers for this Dataset. We search for the network configurations using a parameter range (in the last column), then we choose a final configuration for testing the model (reported in the first columns). Similar to the imputation task, the parameter ranges are chosen as multiples of the number of total features and the number of auxiliary variables $x$, however, sometimes we cap the numbers at 1200 and 800 to prevent too large models.

\begin{table}[ht]
\caption{Description of the imputation model with three linear layers for four forecasting datasets. }
\label{tab:my-table.2}
\begin{center}
\renewcommand{\arraystretch}{1.15}
\begin{tabular}{cm{6em}m{6em}m{6em}m{6em}c}
\toprule[1pt]\midrule[0.3pt]
\multicolumn{1}{l}{} & \multicolumn{5}{c}{\textbf{Spatial learner $\hat{f}$}} \\
\hline
\textbf{Datasets} & Input & Output layer 1 & Output layer 2 & Output & \multicolumn{1}{l}{ Parameter ranges} \\
\hline
 \textbf{Electricity} &  270 & 1200 & 800 & 51 & multiples of 270 \\
 \textbf{Traffic} &  720 & 1200 & 800 & 142 &  multiples of 360 \\
 \textbf{Weather} &  18 & 288 & 288 & 3 & multiples of 18 \\
 \textbf{ILI} &  6 & 96 & 96 & 1 & multiples of 6 \\
 \hline\hline
\multicolumn{1}{l}{} & \multicolumn{5}{c}{\textbf{Temporal learner $\hat{g}$}} \\
\hline
 \textbf{Datasets} & Input & Output layer 1 & Output layer 2 & Output & \multicolumn{1}{l}{ Parameter ranges} \\
 \hline
 \textbf{Electricity} & 96 & 1200 & 800 & 192 & multiples of 321 \\
 \textbf{Traffic} & 96 & 1200 & 800 & 192 &  multiples of 431 \\
 \textbf{Weather} & 96 & 336 & 336 & 192 & multiples of 21 \\
 \textbf{ILI} & 36 & 112 & 112 & 60 & multiples of 7\\
 \midrule[0.3pt]\bottomrule[1pt]
\end{tabular}
\end{center}
\end{table}

\textbf{Additional Plots.} We present additional plots of model predictions compared with ground truth values below. For Exchange dataset and Electricity dataset, we show different time steps and set of features to demonstrate the performance of our framework. For Weather dataset, we plot the predicted values from two models, the first one uses complete data and the second one trains with missing data. 

\begin{figure}[!]
    \centering
    \begin{subfigure}{0.49\linewidth}
    \centering
    \includegraphics[width=\textwidth]{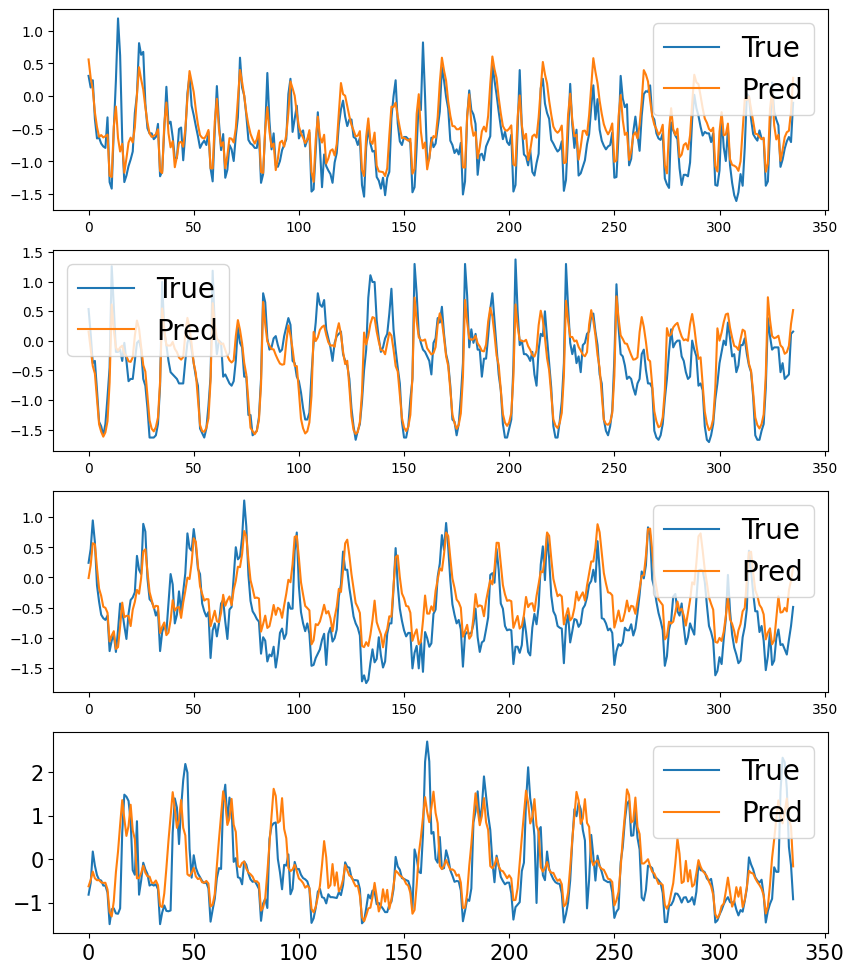}
    \caption{Electricity dataset for features 8,1,4,9}
    \end{subfigure}
    \begin{subfigure}{0.49\linewidth}
    \centering
    \includegraphics[width=\textwidth]{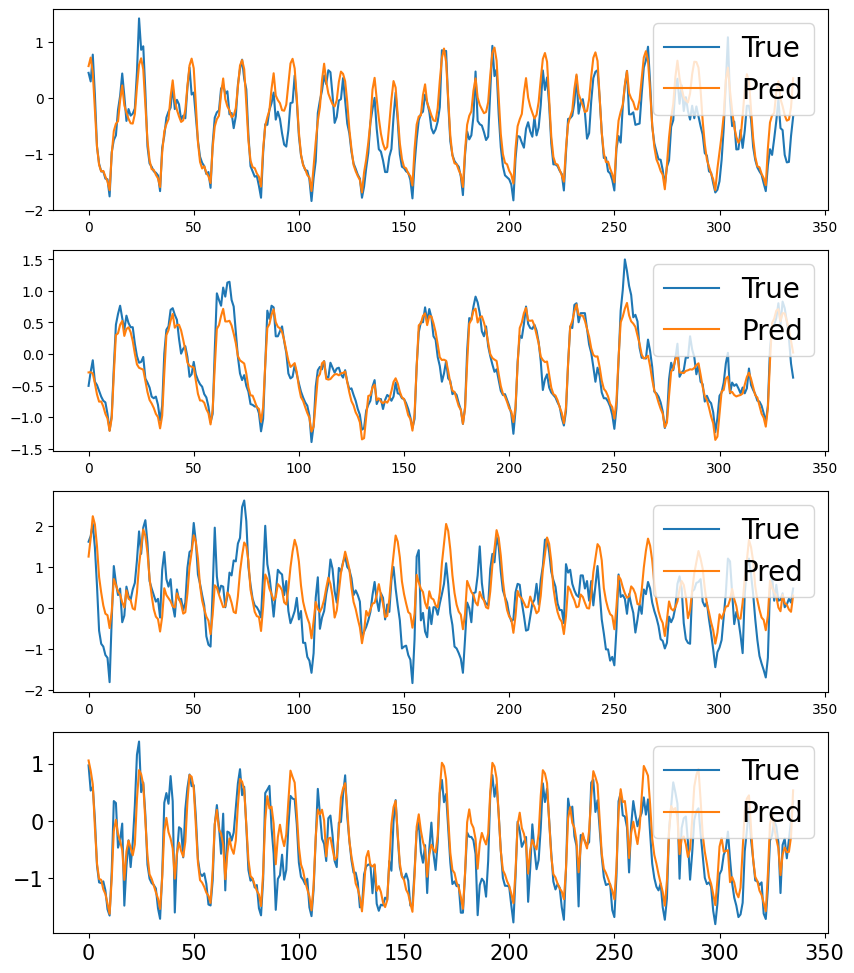}
    \caption{Electricity dataset for features 5,12,17,21}
    \end{subfigure}
    \caption{Predictions from our model for Electricity dataset. Each plot shows the prediction of one feature for 336 future time steps. 
    The Electricity dataset has 321 variables.
    }
    \label{fig:elec}
\end{figure}

\begin{figure}[!]
    \centering
    \begin{subfigure}{0.49\linewidth}
    \centering
    \includegraphics[width=\textwidth]{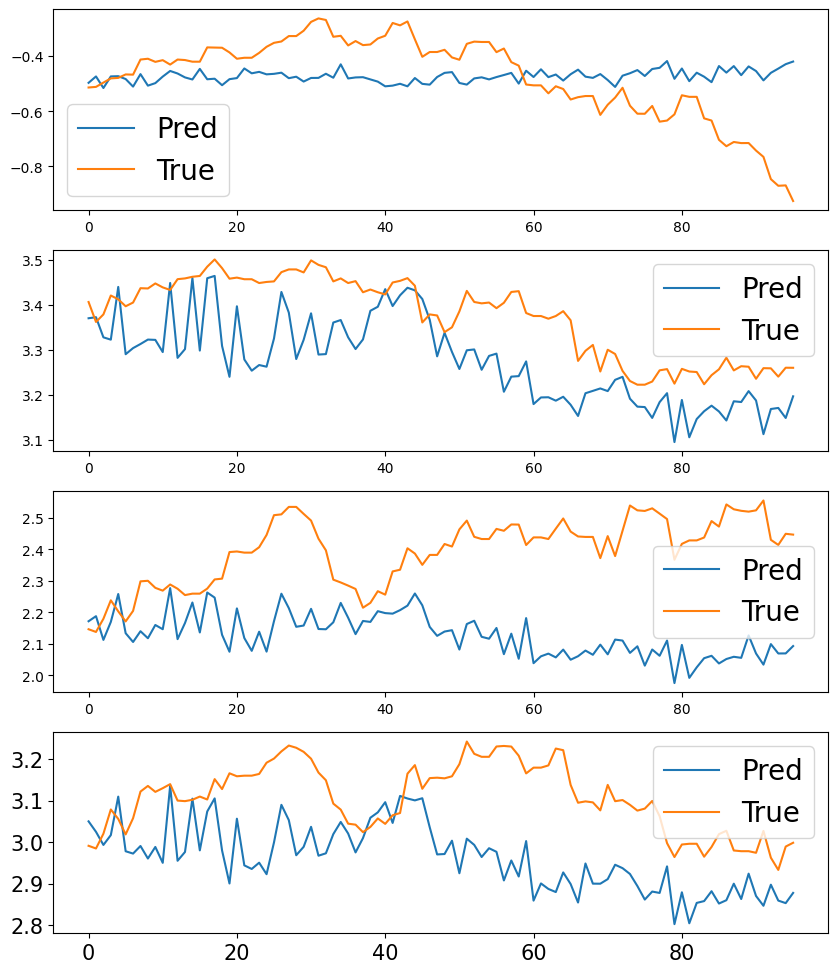}
    \caption{Exchange dataset - 96 future time steps}
    \end{subfigure}
    \begin{subfigure}{0.49\linewidth}
    \centering
    \includegraphics[width=\textwidth]{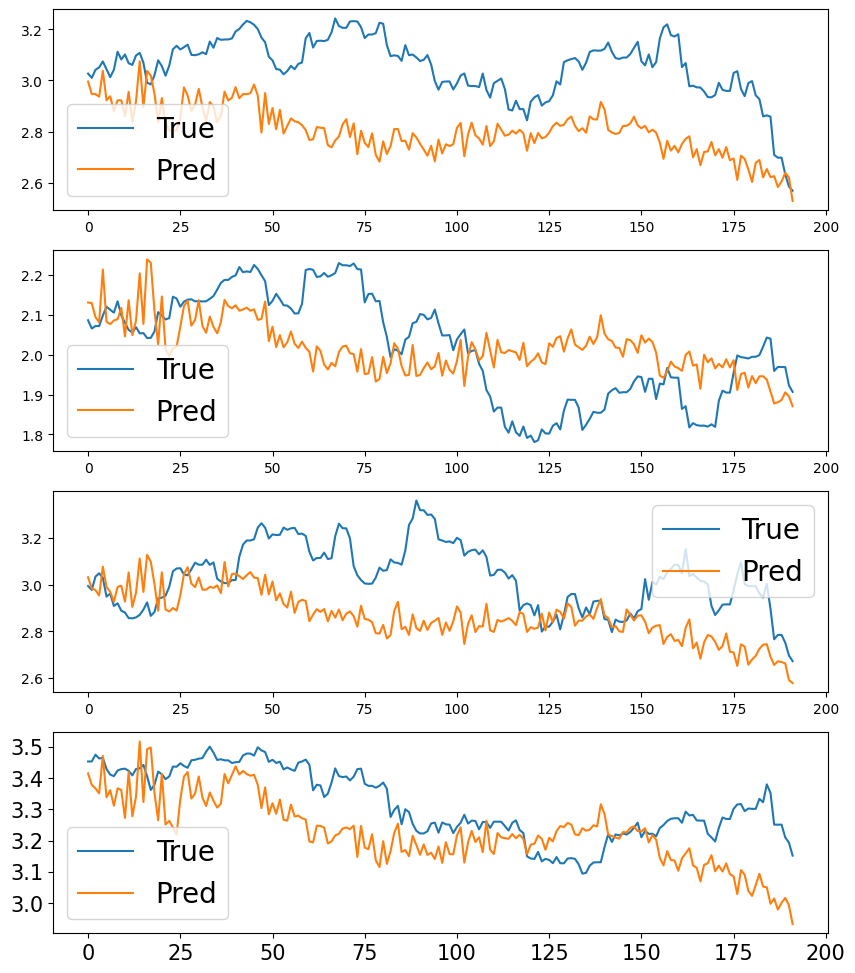}
    \caption{Exchange dataset - 192 future time steps}
    \end{subfigure}
    \caption{Predictions from our model for Exchange dataset for 96 future time steps (in the left) and for 192 future time steps (in the right). Each plot shows the prediction of one feature. 
    The Exchange dataset has 8 variables. 
    }
    \label{fig:exch}
\end{figure}

Figure \ref{fig:elec} shows predictions from our model for Electricity dataset, where each plot shows the prediction of one feature for 336 future time steps. Figure \ref{fig:exch} shows predicted values for Exchange dataset for various input lengths, e.g. 96 time steps and 192 time steps. 

\begin{figure}[!]
    \centering
    \begin{subfigure}{0.49\linewidth}
    \centering
    \includegraphics[width=\textwidth]{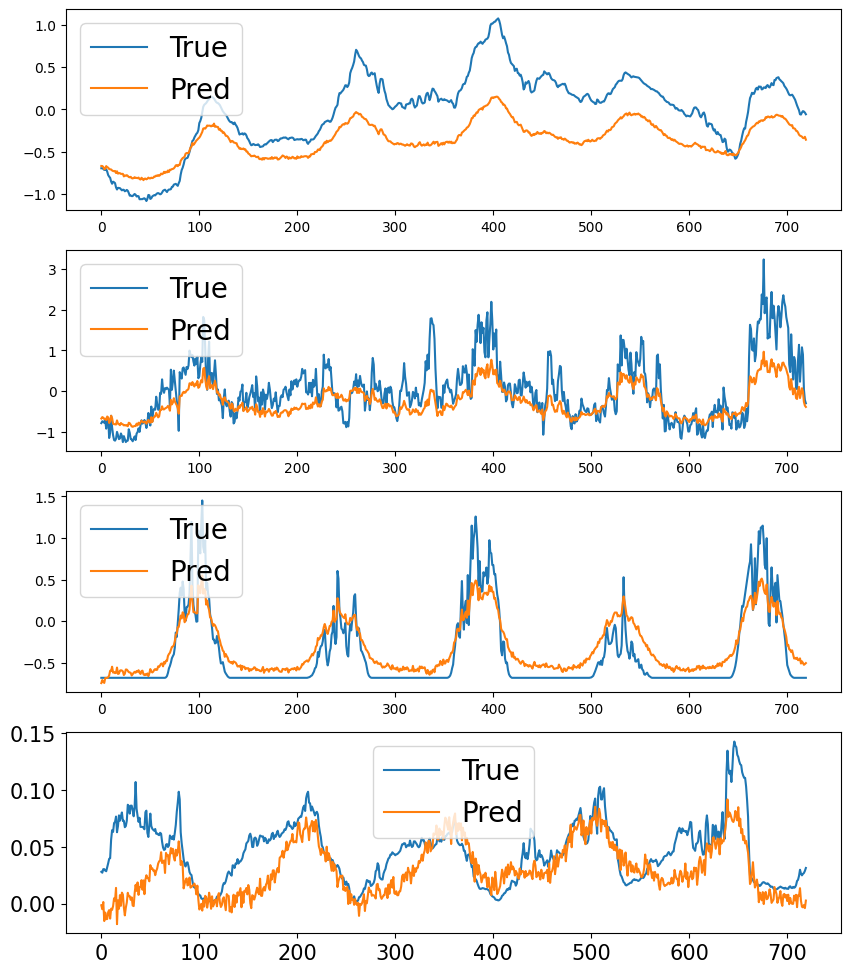}
    \caption{Predictions from model trained by complete data}
    \end{subfigure}
    \begin{subfigure}{0.5\linewidth}
    \centering
    \includegraphics[width=\textwidth]{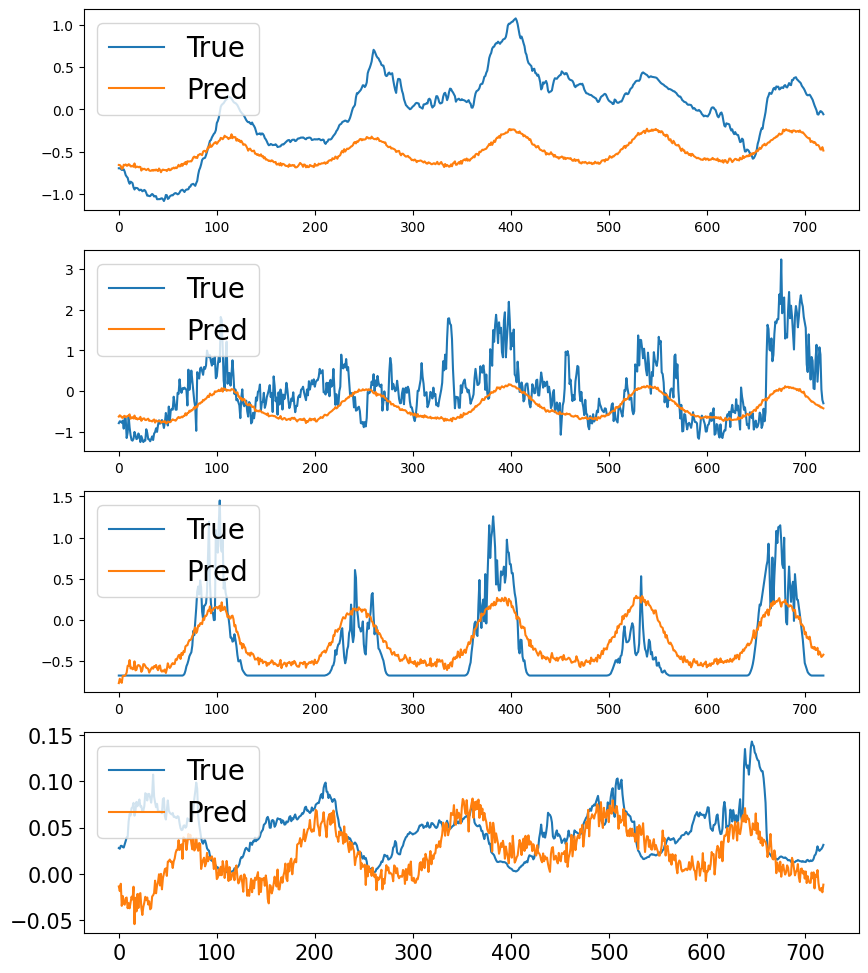}
    \caption{Predictions from model trained by missing data}
    \end{subfigure}
    \caption{Predictions from our model for Weather dataset trained by complete values (in the left) and by missing values (in the right). Each plot shows the prediction of one feature, and we show 720 time steps for Weather data. The Weather dataset has 21 features.}
    \label{fig:weather}
\end{figure}

Finally, Figure \ref{fig:weather} plots the predictions from two models, the first one uses complete data and the second one trains with 20\% level of missing data. From this experiment, we observe that the model with missing data perform slightly worse but still able to capture the dynamic of the time series data. 


\newpage

\newpage

\bibliographystyle{plainnat}
\bibliography{refs}





\end{document}